\documentclass{article}
\usepackage{spconf,amsmath,graphicx}
\usepackage{hyperref}

\usepackage{dblfloatfix}    % To enable figures at the bottom of page
\title{Fully Convolutional neural network for semantic segmentation of anatomical structure and pathologies in colour fundus images associated with diabetic retinopathy }
\name{Oindrila Saha, Rachana Sathish and Debdoot Sheet\thanks{}}
\address{Indian Institute of Technology Kharagpur, West Bengal, India}
\begin{document}
%\ninept
%

\maketitle
\begin{abstract}
Diabetic retinopathy (DR) is the most common form of diabetic eye disease. Retinopathy can affect all diabetic patients and becomes particularly dangerous, increasing the risk of blindness, if it is left untreated. The success rate of its curability solemnly depends on diagnosis at an early stage. The development of automated computer aided disease diagnosis tools could help in faster detection of symptoms with a wider reach and reasonable cost. This paper proposes a method for the automated segmentation of retinal lesions and optic disk in fundus images using a deep fully convolutional neural network for semantic segmentation. This trainable segmentation pipeline consists of an encoder network, a corresponding decoder network followed by pixel-wise classification to segment microaneurysms, hemorrhages, hard exudates, soft exudates, optic disk from background. The network was trained using Binary cross entropy criterion with Sigmoid as the last layer, while during an additional SoftMax layer was used for boosting response of single class. The performance of the proposed method is evaluated using sensitivity, positive prediction value (PPV) and accuracy as the metrices. Further, the position of the Optic disk is localised using the segmented output map.
\end{abstract}


\label{fig:1}
\section{Introduction}
\label{sec:intro}
\begin{figure}[ht]

  \centering
  \centerline{\includegraphics[width=8.5 cm]{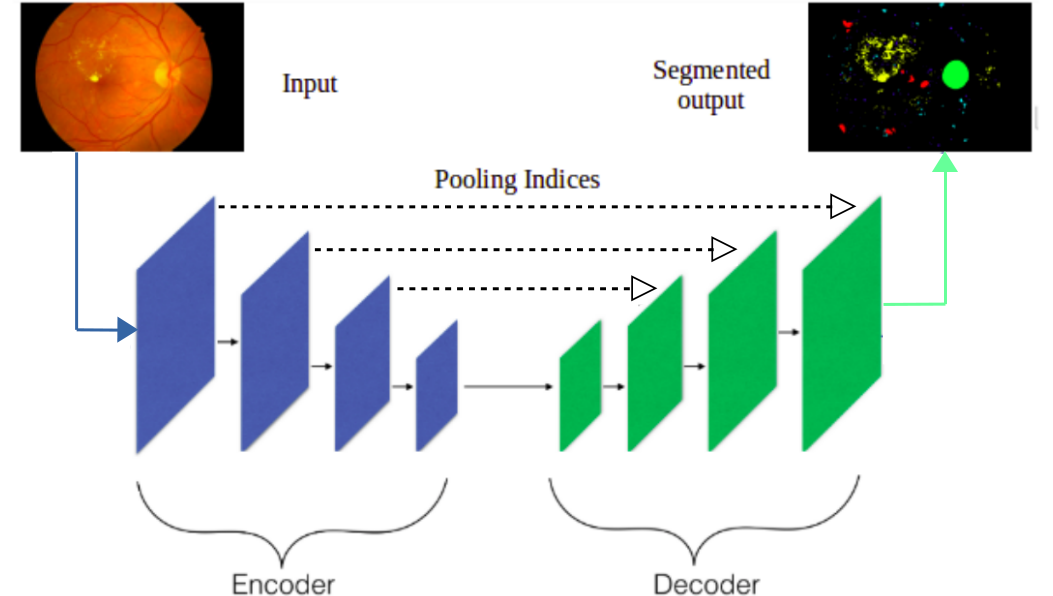}}
%  \vspace{2.0cm}

%
\caption{Overview of proposed method}
\label{fig:2}
\end{figure}

Diabetic retinopathy (DR) is the leading cause of blindness in the working-age population. Screening for DR and monitoring disease progression, especially in the early asymptomatic stages, is effective for preventing visual loss and reducing costs for health systems \cite{dia}. Most screening programs use non-mydriatic digital color fundus cameras to acquire color photographs of the retina. These photographs are then examined for the presence of lesions indicative of DR. The most common signs of DR are red lesions
(microaneurysms, hemorrhages) and bright lesions
(exudates). The presence of red lesions and/or hard exudates (bright lesions) are
indicative of early stage DR. Microaneurysms (MAs) are focal dilatations of retinal capillaries and appear as red
dots in retinal fundus images. Bright lesions or intraretinal
lipid exudates results from the breakdown of blood
retinal barrier. Excluded fluid rich in lipids and proteins
leave the parenchyma, leads to retinal edema and
exudation. Lastly, wherever capillary walls are weak
inside the retina, dot hemorrhages lesions are found which
are slightly larger than MAs. On rupturing it will cause
intra-retinal hemorrhages. Progression of DR also causes
macular edema, neo-vascularization and in later stages,
retinal detachment.

Early methods of semantic segmentation that relied on lowlevel
vision cues have fast been superseded by popular machine
learning algorithms. In particular, deep learning has seen huge success
lately in handwritten digit recognition, speech, categorising
whole images and detecting objects in images \cite{deep}. However, Some
of the recent approaches in semantic pixel-wise labelling using deep CNNs give results which are coarse \cite{coarse}. This is primarily because max pooling and sub-sampling reduce
feature map resolution. SegNet \cite{DBLP:journals/corr/BadrinarayananK15} solves this by mapping low resolution features to input resolution for pixel-wise classification. This mapping produces features
which are useful for accurate boundary localization.
The proposed method uses a end to end trainable segmentation tool consisting of a encoder network followed by a corresponding decoder network and finally a pixel wise classification layer as shown in Fig 1. The architecture of the encoder network is topologically identical to the 13 convolutional layers in the
VGG16 network \cite{vgg} . The role of the decoder network is to map the low resolution encoder feature maps to full input resolution feature
maps for pixel-wise classification. The idea is to capture the global context of the image as the optic disk and hard exudates have similar brightness levels which makes it hard to differentiate them when only the local context is considered.

\section{Methodology}
\label{sec:method}
In the proposed approach, the challenge of segmenting the regions corresponding to microaneurysms, hemorrhages, soft exudates, hard exudates and optic disk is formulated as a task of pixel-wise classification of the retinal images using a fully-convolutional neural network. Optic Disk has been added as a class in the same segmentation problem as lesions, so that the model is better able to differentiate exudates and optic disk. Given an image $\mathcal{I}$ of size M $\times$ N $\times$ 3 the problem of segmentation can be formulated as a pixel-wise classification task where each pixel is assigned a label $\mathcal{L}$ $\in$ \{ ${l_1}$, ${l_2}$, ..${l_C}$ \} such that an output image  $\mathcal{O}$ of size M $\times$ N $\times$ C is generated. Each channel of $\mathcal{O}$ corresponds to one of the classes. In this proposed method a fully convolutional network having an encoder decoder architecture is trained for the task of segmentation using Binary Cross Entropy loss objective function given as 
\begin{equation}
\resizebox{1.0\hsize}{!}
{
$BCE(\mathcal{O}, \mathcal{T}) = - \frac{1}{M \times N} \sum_{i=1}^{M} \sum_{j=1}^{N} (\mathcal{T}_{i,j} \times log(\mathcal{O}_{i,j}) + (1 - \mathcal{T}_{i,j}) \times log(1 - \mathcal{O}_{i,j}))$
}
\end{equation}

\subsection{Network}
SegNet has an encoder network and a corresponding decoder
network, followed by a final pixelwise classification layer. The encoder network consists
of 13 convolutional layers which correspond to the first 13
convolutional layers in the VGG16 network \cite{vgg}. Each encoder layer
has a corresponding decoder layer and hence the decoder network
has 13 layers. In contrast to original SegNet, the final decoder output is fed to a sigmoid layer to produce class probabilities for each pixel
independently in 7 channels. In the target of 7 channels, each channel is the same size as input image : 536 $\times$ 356 and consists of activations in the range [0,1] where 0 corresponds to background and 1 to the presence of corresponding class. 

Each encoder in the encoder network performs convolution
with a filter bank to produce a set of feature maps. These are
then batch normalized \cite{bn}. Then an element-wise rectified linear
non-linearity (ReLU) $max(0, x)$ is applied. Following that,
max-pooling with a 2 $\times$ 2 window and stride 2 (non-overlapping
window) is performed and the resulting output is sub-sampled by
a factor of 2. Max-pooling is used to achieve translation invariance
over small spatial shifts in the input image. Sub-sampling results
in a large input image context (spatial window) for each pixel
in the feature map. While several layers of max-pooling and
sub-sampling can achieve more translation invariance for robust
classification correspondingly there is a loss of spatial resolution
of the feature maps. The increasingly lossy (boundary detail)
image representation is not beneficial for segmentation where
boundary delineation is vital. Therefore, the boundary information in the encoder feature maps are captured and stored before sub-sampling is performed. This is done by storing only the max-pooling indices ( due to memory constraints ),
i.e, the locations of the maximum feature value in each pooling
window is memorized for each encoder feature map.

The appropriate decoder in the decoder network upsamples
its input feature map(s) using the memorized max-pooling indices
from the corresponding encoder feature map(s). This step produces
sparse feature map(s). These feature maps are then convolved with
a trainable decoder filter bank to produce dense feature maps.
A batch normalization step is then applied to each of these maps. The
high dimensional feature representation at the output of the final
decoder is fed to a sigmoid layer. This squashes the value to [0,1] denoting the probabilities for presence of a class for each pixel independently. The output of the sigmoid layer is a 7 channel image of probabilities, each channel denoting one of the classes. 
\subsection{Training}
For training, the images were downsampled to 536 $\times$ 356 which is exactly ${1/8^{th}}$ of original images keeping aspect ratio same. Patch-wise training of the network was not resorted to as the patches containing exudates and optic disk had similar intensity which renderes the task of differentiating them rather difficult.
In addition to the dataset released for the challenge, Drishti-GS \footnote{http://cvit.iiit.ac.in/projects/mip/drishti-gs/mip-dataset2/Home.php} dataset was used for data augmentation. The images of the Drishti-GS dataset \cite{drishti} were resized keeping the aspect ratio intact and zero-padded to bring it to 536 $\times$ 356 pixels. For Optic Disk mask for the Drishti GS dataset, the values $\geq$0.75 were taken from the segmentation soft map which signifies agreement by 3 of 4 experts upon presence of optic disk. Data was further augmented by also taking horizontal, vertical and 180 degree flipped versions of the original images. Two additional classes were introduced, which are the retinal disk excluding the lesions and optic disk; the black background. The retinal disk was found by thresholding a grayscale version of the fundus image.

\begin{figure*}
\centering
   \includegraphics[width=\textwidth,height=8cm]{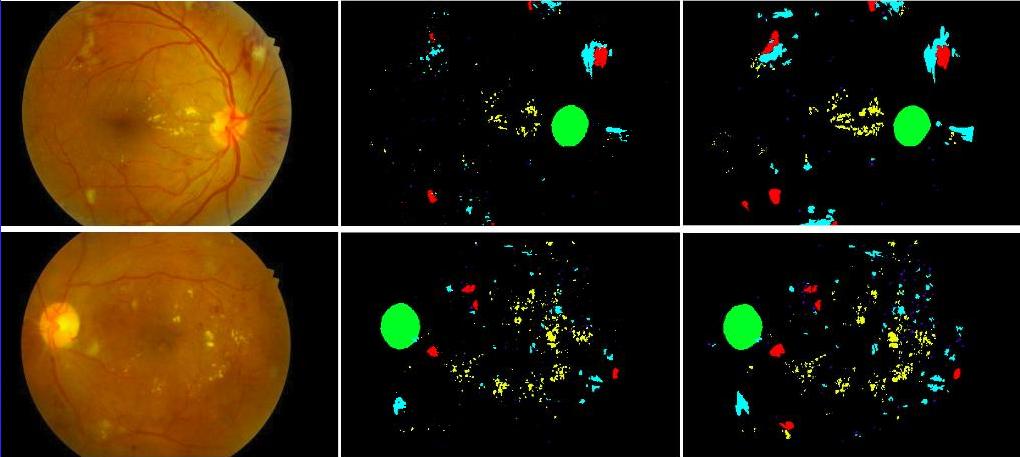}
  \caption{The first column shows the retina fundus images, the second the predicted segmented masks while the third column shows the ground truth segmented masks. Green : Optic Disk, Red: Soft Exudates, Blue: Microaneurysms, Cyan: Haemorrhages, Yellow: Hard Exudates}
\end{figure*}

We use the Binary Cross Entropy loss as the objective function. The losses are averaged for each minibatch over observations as well as over dimensions. It is observed that overfitting happens for different classes at different epochs. By monitoring the validation loss, different stages of the trained network are saved which gives best performance for each of the classes so as to create an ensemble of networks for inference. 
\subsection{Inference}
During inference we introduce an additional Softmax layer after the Sigmoid layer which normalizes the value of a pixel for each class across channels. The Softmax layer has no trainable parameters, hence inclusion in inference is not dependant on training. On using the Softmax layer during inference the masks come closer to groundtruth than the case without Softmax. Finally, segmented ouput is upsampled for each class to 4288 $\times$ 2848 and compared with the groundtruths. Localization of optic disk is done by finding the centroid of the region segmented out as the optic disk which is obtained after thresholding the output of the trained network. 

\section{EXPERIMENTS}
\label{sec:experiments}
The dataset used for this problem is from the IDRiD Diabetic Retinopathy Segmentation and Grading challenge. The dataset for the challenge however provides optic disk segmentation mask for the images with Apparent Retinopathy only. Hence, to identify the presence of optic disk as well as absence of lesions in images with no apparent retinopathy, we used the Drishti-GS dataset. The retinal fundus images of the Drishti-GS dataset, like IDRiD, are all collected from Indians. Importantly, they do not have the presence of any lesions and have the segmented map of optic disk available.

We use Adam optimizer \cite{adam} with learning rate ${10^{-3}}$ and $\beta$ 0.9. Early stopping of the training based on the validation loss is adopted to prevent overfitting. It was observed that the validation loss started to increase after 200 epochs. Before choosing proposed methodology we conducted previous experiments. In the first experiment, the network was trained using patches of size 256 $\times$ 256  extracted from the original image of size 4288 $\times$ 2848 pixels. The second experiment considered only a 5 class problem without including the retinal disk and black background as separate classes.

\section{RESULTS AND DISCUSSIONS}
\label{sec:rnd}
Sensitivity and Positive predictive value (PPV), is plotted against the threshold for each class for obtained grayscale segmented masks. It is observed that as threshold is increased Sensitivity reduces whereas PPV increases. 
\begin{figure}[h]
  \centering
  \centerline{\includegraphics[width=6.5 cm]{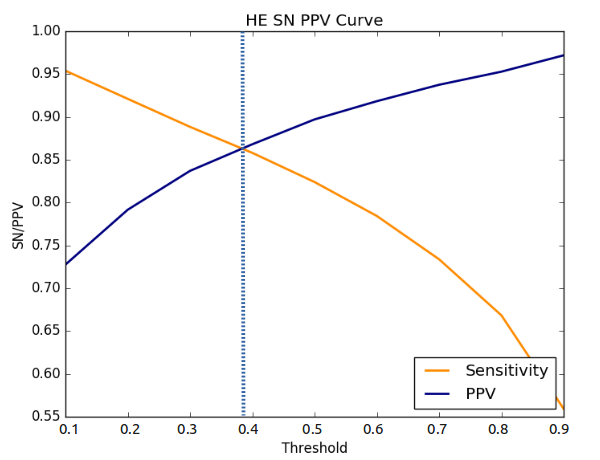}}
\caption{Sensitivity PPV tradeoff}
\label{fig:3}
\end{figure}
The best threshold was then chosen from the point of intersection of these two plots as shown in Figure 3. As shown, for HE - Haemorrhages we choose 0.4 as the best threshold. Similarly, for each class the best threshold was chosen from the intersection points.

The qualitative results are shown in Fig 2. Table 1 lists the quantitative results.
\begin{table}[h]
\begin{center}
  \begin{tabular}{ | c | c | c | c |}
    \hline
    Class & Metric & Score  \\ \hline
    Optic Disk & Jaccard Index (IOU) & 0.8572 \\ \hline
    Microaneurysms & Area under PPV vs SE & 0.0059 \\ \hline
    Hard Exudates & Area under PPV vs SE & 0.5498 \\ \hline
    Haemorrhages & Area under PPV vs SE & 0.0829 \\ \hline
    Soft Exudates & Area under PPV vs SE & 0.1823 \\
    \hline
    
  \end{tabular}
  \caption{Quantitative Results}
\end{center}
\end{table}

Of the previous experiments, on training using patches, the model was not able to distinguish between optic disk and exudates due to lack of global view. Also taking a 7 class problem in place of a 5 class one improves results by a considerable margin.

The evaluation of the localisation of optic disk was done by finding the Euclidean distance between the predicted and the ground truth; after segmenting the optic disk using the above method and finding the centroid. The given ground truth locations of optic disk were not used for training, but the evaluation was done using the provided locations. The mean Euclidean distance for given 413 images was found to be $65.93$ .

\section{CONCLUSION}
\label{sec:conclusion}

Previous work \cite{prevv} in automatic detection of diabetic retinopathy deal with grading or identifying stage of the disease. These methods require extensive pre processing and many steps to finally reach the result \cite{prev}. The proposed method solves the problem of segmentation of lesions in an end-to-end manner. The segmented output can then also be leveraged to determine the severity level directly. Also, no manual feature extraction is needed in our process. This method provides a single unified solution for six of the subtasks. Further work however needs to be done to be better able to identify the Microaneurysm lesions.

\bibliographystyle{IEEEbib}

\begin{thebibliography}{10}

\bibitem{dia}
Martin~M Nentwich and Michael~W Ulbig,
\newblock ``Diabetic retinopathy-ocular complications of diabetes mellitus,''
\newblock {\em World journal of diabetes}, vol. 6, no. 3, pp. 489, 2015.

\bibitem{deep}
Christian Szegedy, Wei Liu, Yangqing Jia, Pierre Sermanet, Scott~E. Reed,
  Dragomir Anguelov, Dumitru Erhan, Vincent Vanhoucke, and Andrew Rabinovich,
\newblock ``Going deeper with convolutions,''
\newblock {\em CoRR}, vol. abs/1409.4842, 2014.

\bibitem{coarse}
Liang{-}Chieh Chen, George Papandreou, Iasonas Kokkinos, Kevin Murphy, and
  Alan~L. Yuille,
\newblock ``Semantic image segmentation with deep convolutional nets and fully
  connected crfs,''
\newblock {\em CoRR}, vol. abs/1412.7062, 2014.

\bibitem{DBLP:journals/corr/BadrinarayananK15}
Vijay Badrinarayanan, Alex Kendall, and Roberto Cipolla,
\newblock ``Segnet: {A} deep convolutional encoder-decoder architecture for
  image segmentation,''
\newblock {\em CoRR}, vol. abs/1511.00561, 2015.

\bibitem{vgg}
Karen Simonyan and Andrew Zisserman,
\newblock ``Very deep convolutional networks for large-scale image
  recognition,''
\newblock {\em CoRR}, vol. abs/1409.1556, 2014.

\bibitem{bn}
Sergey Ioffe and Christian Szegedy,
\newblock ``Batch normalization: Accelerating deep network training by reducing
  internal covariate shift,''
\newblock {\em CoRR}, vol. abs/1502.03167, 2015.

\bibitem{drishti}
J.~Sivaswamy, S.~R. Krishnadas, G.~Datt Joshi, M.~Jain, and A.~U.~Syed Tabish,
\newblock ``Drishti-gs: Retinal image dataset for optic nerve head(onh)
  segmentation,''
\newblock in {\em 2014 IEEE 11th International Symposium on Biomedical Imaging
  (ISBI)}, April 2014, pp. 53--56.

\bibitem{adam}
Diederik~P. Kingma and Jimmy Ba,
\newblock ``Adam: {A} method for stochastic optimization,''
\newblock {\em CoRR}, vol. abs/1412.6980, 2014.

\bibitem{prevv}
Gulshan V, Peng L, Coram M, and et~al,
\newblock ``Development and validation of a deep learning algorithm for
  detection of diabetic retinopathy in retinal fundus photographs,''
\newblock {\em JAMA}, vol. 316, no. 22, pp. 2402--2410, 2016.

\bibitem{prev}
Muthu Rama~Krishnan Mookiah, U.~Rajendra Acharya, Chua~Kuang Chua, Choo~Min
  Lim, E.Y.K. Ng, and Augustinus Laude,
\newblock ``Computer-aided diagnosis of diabetic retinopathy: A review,''
\newblock {\em Computers in Biology and Medicine}, vol. 43, no. 12, pp. 2136 --
  2155, 2013.

\end{thebibliography}

\end{document}